\documentclass[10pt,twocolumn,letterpaper]{article}

\usepackage{cvpr}
\usepackage{times}
\usepackage{epsfig}
\usepackage{graphicx}
\usepackage{amsmath}
\usepackage{amssymb}
\usepackage{siunitx}
\usepackage{balance}
\usepackage{multirow}
\usepackage{hhline}
\usepackage[breaklinks=true,bookmarks=false]{hyperref}

\cvprfinalcopy 

\def\cvprPaperID{7086} 

\begin{document}
\renewcommand\footnotemark{}
\title{Weakly Supervised Video Moment Retrieval From Text Queries}

\author{{Niluthpol Chowdhury Mithun\textsuperscript{*},~ Sujoy Paul}\textsuperscript{*}\thanks{\textsuperscript{*}Joint first author},~ Amit K. Roy-Chowdhury\\
Electrical and Computer Engineering, University of California, Riverside\\
{\tt\small \{nmithun,~supaul,~amitrc\}@ece.ucr.edu}
}

\maketitle


\begin{abstract}
There have been a few recent methods proposed in text to video moment retrieval using natural language queries, but requiring full supervision during training. However, acquiring a large number of training videos with temporal boundary annotations for each text description is extremely time-consuming and often not scalable. In order to cope with this issue, in this work, we introduce the problem of learning from weak labels for the task of text to video moment retrieval. The weak nature of the supervision is because, during training, we only have access to the video-text pairs rather than the temporal extent of the video to which different text descriptions relate. We propose a joint visual-semantic embedding based framework that learns the notion of relevant segments from video using only video-level sentence descriptions. Specifically, our main idea is to utilize latent alignment between video frames and sentence descriptions using Text-Guided Attention (TGA). TGA is then used during the test phase to retrieve relevant moments. Experiments on two benchmark datasets demonstrate that our method achieves comparable performance to state-of-the-art fully supervised approaches.

\end{abstract}


\section{Introduction}

\begin{figure}[t]
\vspace{0.3cm}
\centering
	\includegraphics[width=0.47\textwidth]{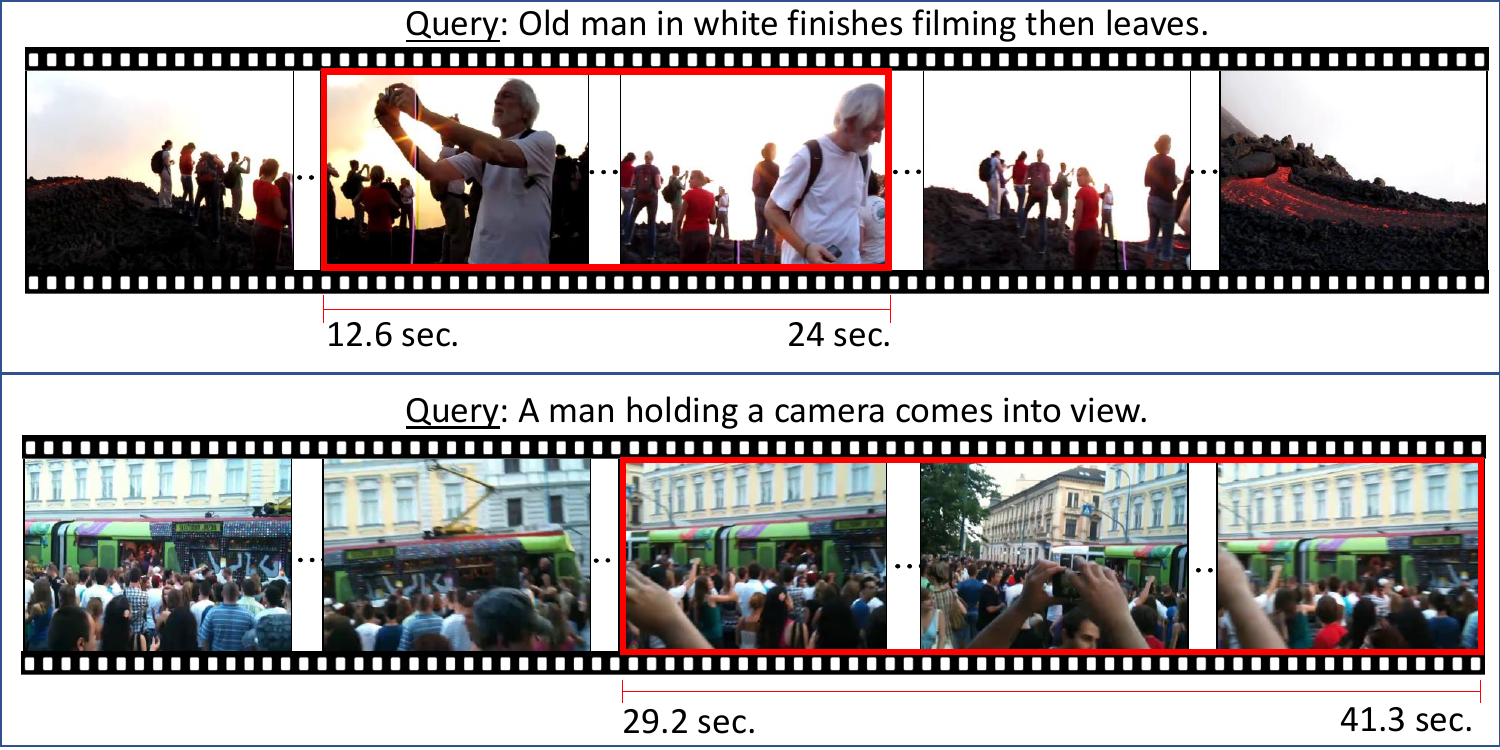}
	\caption{Illustration of text to video moment retrieval task: given a text query, retrieve and rank videos segments based on how well they depict the text description. }
	\label{fig:problem_description}
	\vspace{-0.4cm}
\end{figure}

Cross-modal retrieval of visual data using natural language description has attracted intense attention in recent years \cite{henning2017estimating, zhang2017multi, kiros2014unifying,karpathy2015deep, xu2018text,xu2016msr,nam2016dual}, but remains a very challenging problem~\cite{zhang2017multi,FaghriFKF17,mithun2018learning} due to the differences and ambiguity between different modalities. The identification of the video moment (or segment) is important since it allows the user to focus on the portion of the video that is most relevant to the textual query, and is beneficial when the video has a lot of non-relevant portions. (See Fig.~\ref{fig:problem_description}). The aforementioned approaches operate in a fully supervised setting, i.e., they have access to text descriptions along with the exact temporal location of the visual data corresponding to the descriptions. However, obtaining such annotations is tedious and noisy, requiring multiple annotators. The process of developing algorithms which demand a weaker degree of supervision is non-trivial and is yet to be explored by researchers for the problem of video moment retrieval using text queries. In this work, we focus, particularly on this problem.  

\begin{figure*}[t]
\centering
	\includegraphics[width=0.98\textwidth]{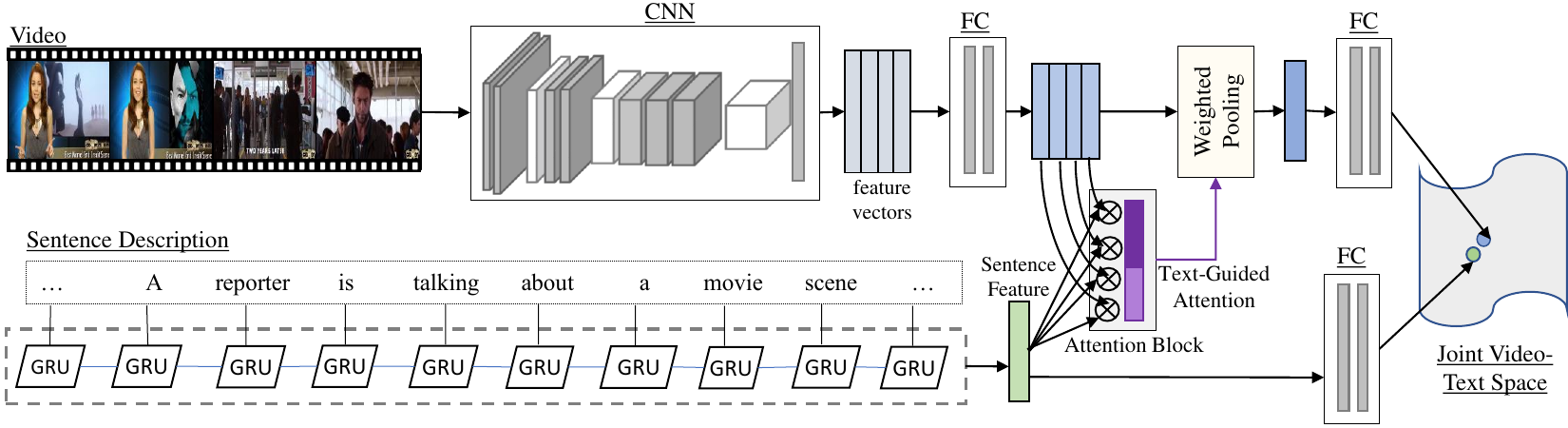}
	\vspace{-0.1cm}
	\caption{A brief illustration of our proposed weakly supervised framework for learning joint embedding model with Text-Guided Attention for text to video moment retrieval. Our framework learns a latent alignment between relevant video frames and text corresponding to the video. This alignment is utilized for attending video features based on relevance and the pooled video feature is used for learning the joint video-text embedding. In the figure, CNN refers to a convolutional neural network, FC refers to a fully-connected neural network, and GRU refers to gated recurrent units. Please see Sec.~\ref{methods} for details of our approach.}
	\vspace{-0.3cm}
	\label{fig:block_diagram}
\end{figure*}

The text to video moment retrieval task is more challenging than the task of localizing categorical activities in videos, which is a comparatively well-studied field \cite{ma2016learning,wang2017untrimmednets,zhao2017temporal,xu2017r,paul2018w,singh2016multi}. Although these methods show success on activity localization, unlike text to moment retrieval, they are limited to a pre-defined set of activity classes. In this regard, there has been a recent interest in localizing moments in a video from natural language description \cite{hendricks2017localizing, gao2017tall, xu2018text, chen2018temporally}. Supervision in terms of text description with their temporal boundaries in a video is used to train these models. However, acquiring such dense annotations of text-temporal boundary tuples are often tedious, as it is difficult to mark the start and end locations of a certain moment, which introduces ambiguity in the training data.

On the contrary, it is often much easier to just describe the moments appearing in a video with a set of natural language sentences, than providing exact temporal boundaries associated with each of the sentences. Moreover, such descriptions can often be obtained easily from captions through some sources on the web. Motivated by this, we pose a question in this paper:  \textit{is it possible to develop a weakly-supervised framework for video moment localization from the text, leveraging only video-level textual annotation, without their temporal boundaries?} Temporal localization of moments using weak description is a much more challenging task than its supervised counterpart. It is extremely relevant to address this question, due to the difficulty and non-scalability of acquiring precise frame-wise information with text descriptions in the fully supervised setting, which require enormous manual labor. 

\vspace{0.05cm}
\noindent \textbf{Overview of the Proposed Framework.} An illustration of our proposed weakly-supervised framework presented in Fig.~\ref{fig:block_diagram}. Given a video, we first extract frame-wise visual features from pre-trained Convolutional Neural Network (CNN) architectures. We also extract features for text descriptions using Recurrent Neural Network (RNN) based models. 
Similar to several cross-modal video-text retrieval models \cite{dong2016word2visualvec, kiros2014unifying}, we train a joint embedding network to project video features and text features into the same joint space. However, as we have text descriptions for the videos as a whole and not moment-wise descriptions like in a fully supervised setting, the learning procedure for text to video moment retrieval is non-trivial. 

Given a certain text description, we obtain its similarity with the video features, which gives an indication of temporal locations which may correspond to the textual description. We call this Text-Guided Attention as it helps to highlight the relevant temporal locations, given a text description. Thereafter, we use this attention to pool the video features along the temporal direction to obtain a single text-dependent feature vector for a video. We then train the network to minimize a loss which reduces the distance between the text-dependent video feature vector and the text vector itself. We hypothesize that along with learning a shared video-text embedding, hidden units will emerge internally to learn the notion of relevance between moments of video and corresponding text description. During the testing phase, we use TGA for localizing the moments, given a text query, as it highlights the portion of the video corresponding to the query. 



\vspace{0.1cm}
\underline{\textit{Contributions: }} The main contributions of the proposed approach are as follows.

$\bullet$ We address a novel and practical problem of temporally localizing video moments from text queries without requiring temporal boundary annotations of the text descriptions while training but using only the video-level text descriptions.

$\bullet$ We propose a joint visual-semantic embedding framework, that learns the notion of relevant moments from video using only video-level description. Our joint embedding network utilizes latent alignment between video frames and sentence description as Text-Guided Attention for the videos to learn the embedding.

$\bullet$ Experiments on two benchmark datasets: DiDeMo \cite{hendricks2017localizing} and Charades-STA \cite{gao2017tall} show that our weakly-supervised approach performs reasonably well compared to supervised baselines in the task of text to video moment retrieval.

\section{Related Works}

\textbf{Image/Video Retrieval using Text Queries.} Cross-modal language-vision retrieval methods focus on retrieving relevant images/videos from a database given text descriptions. Most of the recent methods for image-text retrieval task focus on learning joint visual-semantic embedding models \cite{karpathy2014deep,kiros2014unifying,frome2013devise,wang2017learning,FaghriFKF17,nam2016dual,vendrov2015order,mithun2018webly}. Inspired by the success of these approaches, most video-text retrieval methods also employ a joint subspace model \cite{xu2015jointly,dong2016word2visualvec,venugopalan2015sequence,pan2016jointly,mithun2018learning,mithun2019joint}. In this joint space, the similarity of different points reflects the semantic closeness between their corresponding original inputs. These text-based video retrieval approaches focus on retrieving an entire video from dataset given text description.  However, we focus on temporally localizing a specific moment relevant to a text query, within a given video. Similar to the video/image to text retrieval approaches, our proposed framework is also based on learning joint video-text embedding models. However, instead of focusing only on aligning video and text in the joint space as in video-text retrieval, our aim is to learn a latent alignment between video frames and text descriptions, which is used for obtaining the relevant moments corresponding to a given text query.


\vspace{0.1cm}
\textbf{Activity Localization.}
The moment retrieval aspect of our work is related to the problem of temporal activity localization in untrimmed videos. From the perspective of our interest, the works in literature pertaining to activity localization can be categorized as either fully supervised or weakly supervised. Works in fully supervised setting include SSN \cite{zhao2017temporal}, R-C3D \cite{xu2017r}, TAL-Net \cite{chao2018rethinking} among others. Most of these works structure their framework by using temporal action proposals with activity location predictors. However, in the weakly supervised setting, the exact location of each activity is unknown, and only the video-level labels are accessible during training. In order to deal with that, researchers take a Multiple Instance Learning approach \cite{wang2017untrimmednets} with constraints applied for better localization \cite{paul2018w,nguyen2018weakly}. Our task of video moment retrieval from text description is more challenging than the activity localization task, as our method is not limited to a pre-defined set of categories, but rather sentences in natural language. 

\vspace{0.1cm}
\textbf{Text to Video Moment Retrieval.}
Most relevant to our work are the methods that focus on identifying relevant portions from text description using fully-supervised annotations: MCN \cite{hendricks2017localizing}, CTRL \cite{gao2017tall}, EFRC \cite{xu2018text}, ROLE \cite{liu2018cross}, TGN \cite{chen2018temporally}. These methods are severely plagued by the issue of collecting training videos with temporal natural language annotation. Temporal sliding window over videos frames \cite{hendricks2017localizing}, or hard-coded segments containing a fixed number of frames \cite{gao2017tall} has been used for generating moment candidate corresponding to a text description. Moreover, unlike in images, generating temporal proposals for videos in an unsupervised manner is itself a challenging task. In \cite{xu2018text,xu2017r}, the authors proposed an end-to-end framework where the activity proposals are generated as one of the initial steps, but for the much easier task of activity localization. Attention mechanism has been used in \cite{liu2018cross, xu2018text} for the text to video moment retrieval task. Although we also use attention, our usage is significantly different from them. ROLE \cite{liu2018cross} uses attention over the words using video moment context, which they obtain from the temporal labels. EFRC \cite{xu2018text} uses attention in training a temporal proposal network as it has access to temporal boundary annotations of the sentences. We use attention over the temporal dimension of the videos as we do not have access to the temporal boundaries. More importantly, our method is weakly-supervised, which requires only video-level text annotation during training. Hence, the data collection cost for our approach is substantially less, and it is possible to acquire and train using larger video-text captioning datasets.

A weakly supervised setting is considered in \cite{bojanowski2015weakly} for the video-text alignment task, which is to assign temporal boundaries to a set of temporally ordered sentences, whereas our task is to retrieve a portion of the video given a sentence. Moreover, \cite{bojanowski2015weakly} assumes temporal ordering between the sentences as additional supervision. Also, their method would require dense sentence annotations describing all portions of the video including tokens representing background moments (if any). 
The task considered in this work is a generalization of the task in~\cite{bojanowski2015weakly}. We consider that there can be multiple sentences describing different temporal portions of a single video and do not consider any temporal ordering information of the sentences. The Text-Guided Attention mechanism used in our framework allows us to deal with multiple sentence descriptions during training and provide the relevant portions for each of them during testing, even with weak supervision.


\section{Approach} \label{methods}
In this section, we first describe the network architecture and input feature representation for representing video and text (Sec.~\ref{app:feat}). Then, we present our proposed Text-Guided Attention module (Sec.~\ref{app:attn}). Finally, we describe the framework for learning joint video-text embedding (Sec.\ref{app:alignment_objective}).

\vspace{0.1cm}
\textbf{Problem Definition.} In this paper, we consider that the training set consists of videos paired with text descriptions composed of multiple sentences. Each sentence describes different temporal regions of the video. However, we do not have access to the temporal boundaries of the moments referred to by the sentences. At test time, we use a sentence to retrieve relevant portions of the video.

\subsection{Network Structure and Features} \label{app:feat}

\textbf{Network Structure.} The joint embedding model is trained using a two-branch deep neural network model, as shown in Fig.~\ref{fig:block_diagram}. The two branches consist of different expert neural networks to extract modality-specific representations from the given input. The expert networks are followed by fully connected embedding layers which focus on transforming the modality-specific representations to joint representations. In this work, we keep the pre-trained image encoder fixed as we have limited training data. The fully-connected embedding layers, the word embedding, the GRU are trained end-to-end. We set the dimensionality ($D$) of the joint embedding space to 1024.

\vspace{0.1cm}
\textbf{Text Representation.} We use Gated Recurrent Units (GRU) \cite{chung2014empirical} for encoding the sentences. GRU has been very popular for generating a representation for sentences in recent works \cite{FaghriFKF17, kiros2014unifying}. The word embeddings are input to the GRU. The dimensionality of the word embeddings is 300.

\vspace{0.1cm}
\textbf{Video Representation.} We utilize pre-trained convolutional neural network models as the expert network for encoding videos. Specifically, following \cite{gao2017tall} we utilize C3D model \cite{tran2015learning} for feature extraction from every 16 frames of video for the Charades-STA dataset. A 16 layer VGG model \cite{simonyan2014very} is used for frame-level feature extraction in experiments on DiDeMo dataset following \cite{hendricks2017localizing}. We extract features from the penultimate fully connected layer. For both the C3D and VGG16 model, the dimension of the representation from the penultimate fully connected layer is 4096.

\begin{figure*}[t]
\centering
	\includegraphics[width=0.78\textwidth]{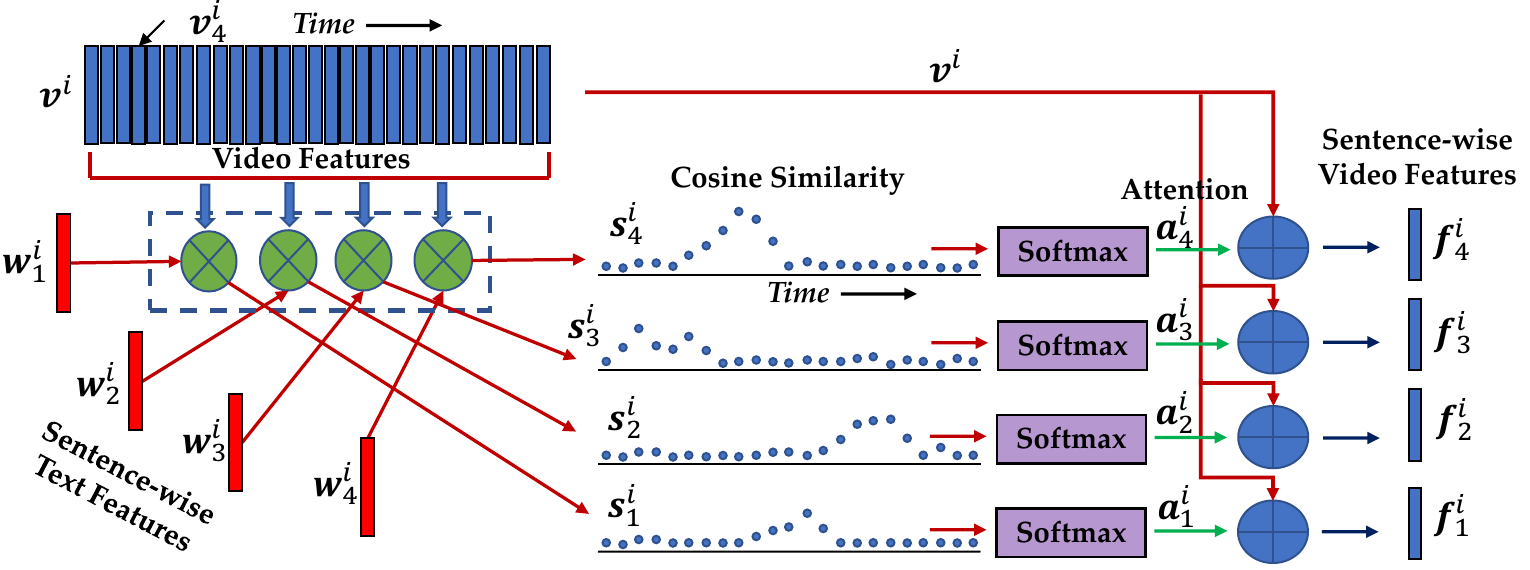}
	\caption{This figure presents the procedure of computing the Text-Guided Attention and using it to generate sentence-wise video features. We first obtain the cosine similarity between the features at every time instant of the video $\boldsymbol{v}_i$, and its corresponding sentences $\boldsymbol{w}^i_j$, followed by a softmax layer along the temporal dimension to obtain the sentence-wise temporal attention. Thereafter, we use these attentions to compute a weighted average of the video features to finally obtain the sentence-wise video features.}
	\vspace{-0.4cm}
	\label{fig:tga}
\end{figure*}

\vspace{-0.1cm}
\subsection{Text-Guided Attention} \label{app:attn}
\vspace{-0.05cm}
After the feature extraction process, we have a training set $\mathcal{D}=\{\{\boldsymbol{w}^i_j\}_{j=1}^{nw_i}, \{\boldsymbol{v}^i_k\}_{k=1}^{nv_i}\}_{i=1}^{n_d}$, where $n_d$ is the number of training pairs, $\boldsymbol{w}^i_j$ represents the $j^{th}$ sentence feature of $i^{th}$ video, $\boldsymbol{v}^i_k$ represent the video feature at the $k^{th}$ time instant of the $i^{th}$ video, $nw_i$ and $nv_i$ are the number of sentences in the text description and video time instants for the $i^{th}$ video in the dataset. Please note that we do not consider any ordering in the text descriptions.

Each of the sentences provides us information about a certain part of the given video. In a fully supervised setting, where we have access to the temporal boundaries associated with each sentence, we can apply a pooling technique to first pool the relevant portion of the video features and then use a similarity measure to learn a joint video segment-text embedding. However, in our case of weakly supervised moment retrieval, we do not have access to the temporal boundaries associated with the sentences. Thus, we need to first obtain the portions of the video which are relevant to a given sentence query. 

If some portion of the video frames corresponds to a particular sentence, we would expect them to have similar features. Thus, the cosine similarity between text and video features should be higher in the temporally relevant portions and low in the irrelevant ones. Moreover, as the sentence described a part of the video rather than individual temporal segments, the video feature obtained after pooling the relevant portions should be very similar to the sentence description feature. We employ this idea to learn the joint video-text embedding via an attention mechanism based on the sentence descriptions, which we name Text-Guided Attention (TGA). Note that during the test phase, we use TGA to obtain the localization. 

We first apply a Fully Connected (FC) layer with ReLU \cite{krizhevsky2012imagenet} and Dropout \cite{srivastava2014dropout} on the video features at each time instance to transform them into the same dimensional space as the text features. We denote these features as $\boldsymbol{\bar{v}}_k^i$. In order to obtain the sentence specific attention over the temporal dimension, we first obtain the cosine similarity between each temporal feature and sentence descriptions. The similarity  between the $j^{th}$ sentence and the $k^{th}$ temporal feature of the $i^{th}$ training video can be represented as follows,
\begin{equation}
    s^i_{kj}=\frac{{\boldsymbol{w}^i_j}^T\boldsymbol{v}^i_k}{||\boldsymbol{w}^i_j||_2||\boldsymbol{v}^i_k||_2}
    \label{sim}
\end{equation}
Once we obtain the similarity  values for all the temporal locations, we apply a softmax operation along the temporal dimension to obtain an attention vector for the $i^{th}$ video as follows,

\begin{equation}
    a^i_{kj} = \frac{\exp(s^i_{kj})}{\sum_{k=1}^{nv_i} \exp(s^i_{kj})}
    \label{atn}
\end{equation}

These should have high values at temporal locations which are relevant to the given sentence vector $\boldsymbol{w}^i_j$. We consider this as local similarity because the individual temporal features may correspond to different aspects of a sentence and thus each of the temporal features might be a bit scattered away from the sentence feature. However, the feature obtained after pooling the video temporal features corresponding to the relevant locations should be quite similar to the entire sentence feature. We consider this global similarity. We use the attention in Eqn. \ref{atn} to obtain the pooled video feature for the sentence description $\boldsymbol{w}^i_j$ as follows,
\begin{equation}
    \boldsymbol{f}^i_j=\sum_{k=1}^{nv_i}a^i_{kj}\boldsymbol{v}^i_{k}
    \label{pool}
\end{equation}

Note that, this feature vector corresponds to the particular sentence description $\boldsymbol{w}^i_j$ only. In a similar procedure, we can extract the text-specific video feature vector corresponding to the other sentences in the text descriptions of the same video and other videos as well. Fig. \ref{fig:tga} presents an overview of the sentence-wise video feature extraction procedure using the video temporal features and a set of sentence descriptions for the video. We use these feature vectors to derive the loss function to be optimized to learn the parameters of the network. This is described next. 

\subsection{Training Joint Embedding} 
\label{app:alignment_objective}
We now describe the loss function we optimize to learn the joint video-text embedding. Many prior approaches have utilized pairwise ranking loss as the objective for learning joint embedding between visual and textual input \cite{kiros2014unifying, zheng2017dual, wang2016learning, karpathy2014deep}. Specifically, these approaches minimize a hinge-based triplet ranking loss in order to maximize the similarity between an image embedding and corresponding text embedding and minimize similarity to all other non-matching ones.

For the sake of notational simplicity, we drop the index $i,j,k$ denoting the video number, sentence index and time instant. Given a text-specific video feature vector based on TGA, $\boldsymbol{f}$ ~($\in \mathbb{R}^V$) and paired text feature vector $\boldsymbol{w}$ ($\in \mathbb{R}^T$), the projection for the video feature on the joint space can be derived as $\boldsymbol{v}_p = W^{(v)}\boldsymbol{f}$ ($\boldsymbol{v}_p \in \mathbb{R}^D$). Similarly, the projection of paired text vector in the embedding space can be expressed as $\boldsymbol{t}_p = W^{(t)}\boldsymbol{w} (\boldsymbol{t}_p \in \mathbb{R}^D)$. Here, $W^{(v)} \in \mathbb{R}^{D \times V}$ is the transformation matrix that projects the video content into the joint embedding and $D$ is the dimensionality of the joint space. Similarly, $W^{(t)}\in \mathbb{R}^{D \times T}$ maps input sentence/caption embedding to the joint space. 

Using these pairs of feature representation of both videos and corresponding sentence, the goal is to learn a joint embedding such that the positive pairs are closer than the negative pairs in the feature space.  Now, the video-text loss function $\mathcal{L}_{VT}$ can be expressed as follows, 
\begin{equation} \label{eq:equation1}
\small
\vspace{-0.3cm}
\begin{split}
\mathcal{L}_{VT} = & \sum_{(\boldsymbol{v}_p,\boldsymbol{t}_p)}\Big\{\sum_{\boldsymbol{t}_p^{-}}max \big[0, \Delta-S(\boldsymbol{v}_p,\boldsymbol{t}_p)+S(\boldsymbol{v}_p,\boldsymbol{t}^{-}_p)\big] \\
 & \hspace{0.4cm} + \ \sum_{\boldsymbol{v}^{-}_p} max \big[0, \Delta-S(\boldsymbol{t}_p,\boldsymbol{v}_p)+S(\boldsymbol{t}_p,\boldsymbol{v}^{-}_p)\big]\Big\}
\end{split}
\vspace{-0.3cm}
\end{equation}
where $\boldsymbol{t}^{-}_p$ is a non-matching text embedding for video embedding $\boldsymbol{v}_p$, and $\boldsymbol{t}_p$ is the matching text embedding. This is similar for video embedding $\boldsymbol{v}_p$ and non-matching image embedding $\boldsymbol{v}^{-}_p$. $\Delta$ is the margin value for the ranking loss. The scoring function $S(\boldsymbol{v}_p,\boldsymbol{t}_p)$ measures the similarity between the image embedding and text embedding in the joint space. We utilize cosine similarity in the representation space to compute similarity. Cosine similarity is widely used in learning joint embedding models in prior works on image-text retrieval \cite{zheng2017dual,kiros2014unifying,FaghriFKF17,mithun2018webly}. Our approach does not depend on any specific choice of similarity function.

In Eq.~(\ref{eq:equation1}), the first term attempts to ensure that for each visual input, the matching text inputs should be closer than non-matching text inputs in learning the joint space. However, the second term in Eq.~(\ref{eq:equation1}) attempts to ensure that for each text input, the matching image input should be closer in the joint space than the non-matching image inputs. 

\subsection{Batch-wise Training}
We train our network using Stochastic Gradient Descent (SGD) by dividing the dataset into batches. For a video with multiple sentences, we create multiple video-sentence pairs, with the same video, but different sentences in the corresponding video's text description. During training, our method learns to automatically identify the relevant portions for each sentence using the Text-Guided Attention.  The negative instances $\boldsymbol{v}_p^-$ and $\boldsymbol{t}_p^-$ correspond to all the instances which are not positive in the current batch of data. 

\section{Experiments} \label{sec:experiments}
\vspace{-0.1cm}

We perform experiments on two benchmark datasets with the goal of comparing the performance of our weakly-supervised approach against different supervised baselines. As we introduce the problem in this paper, to the best of our knowledge, our work is the first to show results on this task. Ideally, any weakly supervised methods would attempt at attaining the performance of the supervised methods, with similar features and setting.

We first describe the details on the datasets and evaluation metric in Sec.~\ref{sec:data}, followed by the training details in Sec.~\ref{sec:training}. Then, we report the results of different methods on DiDeMo and Charades-STA dataset in Sec.~\ref{sec:quantitative}. 

\renewcommand{\arraystretch}{1.1}
\begin{table*}[]
\centering
\caption{This table presents the results on the Charades-STA dataset, using the evaluation protocol in \cite{gao2017tall}. Following \cite{xu2018text,gao2017tall}, we also use C3D feature for a fair comparison. The proposed \textbf{weakly-supervised approach} performs significantly better that visual-semantic embedding based baselines: VSA-RNN and VSA-STV. The proposed approach also performs reasonably comparable to state-of-the-art approaches CTRL\cite{gao2017tall} and EFRC~\cite{xu2018text}, and even achieves a similar performance in some evaluation metrics. }
\vspace{0.1cm}
\small
\begin{tabular}{l||ccc|ccc|ccc}
\hline
\multirow{2}{*}{Method}   & & \underline{IoU=0.3}     &       &  & \underline{IoU=0.5}      &       & & \underline{IoU=0.7}      &       \\ 
         & R@1     & R@5   & R@10  & R@1     & R@5   & R@10  & R@1     & R@5   & R@10  \\  \hhline{=||===|===|===}
Random   & -       & -     & -     & 8.51    & 37.12 & -     & 3.03    & 14.06 & -     \\
VSA-RNN  & -       & -     & -     & 10.50    & 48.43 & -     & 4.32    & 20.21 & -     \\
VSA-STV  & -       & -     & -     & 16.91   & 53.89 & -     & 5.81    & 23.58 & -     \\
CTRL     & -       & -     & -     & 23.63   & 58.92 & -     & 8.89    & 29.52 & -     \\
EFRC     & 53.00      & 94.60    & 98.50  & 33.80    & 77.30  & 91.60  & 15.00      & 43.90  & 60.90  \\
\hline
\textbf{Proposed} & \textbf{29.68}   & \textbf{83.87} & \textbf{98.41} & \textbf{17.04}   & \textbf{58.17} & \textbf{83.44} & \textbf{6.93}    & \textbf{26.80} & \textbf{44.06} \\ \hline
\end{tabular}
\vspace{-0.35cm}
\label{tab:charades}
\end{table*}

\subsection{Datasets and Evaluation Metric}
\label{sec:data}

We present experiments on two benchmark datasets for sentence description based video moment localization, namely Charades-STA \cite{gao2017tall} and DiDeMo \cite{hendricks2017localizing} to evaluate the performance of our proposed framework.

\vspace{0.1cm}
\textbf{Charades-STA.} The Charades-STA dataset for text to video moment retrieval was introduced in \cite{gao2017tall}. The dataset contains 16,128 sentence-moment pairs with 12,408 in the training set and 3,720 in the testing set. The Charades dataset was originally introduced in \cite{sigurdsson2016hollywood} which contains temporal activity annotation and video-level paragraph description for the videos. The authors of \cite{gao2017tall} enhanced the dataset \cite{sigurdsson2016hollywood} for evaluating temporal localization of moments in videos given text queries. The video-level descriptions from the original dataset were decomposed into short sentences. Then, these sentences are assigned to segments in videos based on matching keywords for activity categories. The annotations are manually verified at last.

\vspace{0.1cm}
\textbf{DiDeMo.} The  Distinct Describable Moments (DiDeMo) dataset \cite{hendricks2017localizing} is one of the largest and most diverse datasets for the temporal localization of events in videos given natural language descriptions. The videos are collected from Flickr and each video is trimmed to a maximum of 30 seconds. The videos in the dataset are divided into 5-second segments to reduce the complexity of annotation. The dataset is split into training, validation and test sets containing 8,395, 1,065 and 1,004 videos respectively. The dataset contains a total of 26,892 moments and one moment could be associated with descriptions from multiple annotators. The descriptions in DiDeMo dataset are detailed and contain camera movement, temporal transition indicators, and activities. Moreover, the descriptions in DiDeMo are verified so that each description refers to a single moment.

\vspace{0.1cm}
\textbf{Evaluation Metric.} We use the evaluation criteria following prior works in literature \cite{hendricks2017localizing,gao2017tall}. Specifically, we follow \cite{hendricks2017localizing} for evaluating DiDeMo dataset and \cite{gao2017tall} for evaluating Charades-STA. We measure rank-based performance $R@K$ (Recall at $K$) which calculates the percentage of test samples for which the correct result is found in the top-$K$ retrievals to the query sample. We report results for R@1, R@5, and R@10.  We also calculate temporal intersection over union (tIoU) for Charades-STA dataset and mean intersection over union (mIoU) for DiDeMo dataset. 

\subsection{Implementation Details} \label{sec:training}

We used two Telsa K80 GPUs and implemented the network using PyTorch \cite{paszke2017automatic}. We start training with a learning rate of 0.001 and keep the learning rate fixed for 15 epochs. The learning rate is lowered by a factor of 10 every 15 epochs. We tried different values for margin $\alpha$ in training and found $0.1 \leq \Delta  \leq 0.2$ works reasonably well. We empirically choose $\Delta$ as 0.1 for Charades-STA and 0.2 for DiDeMo in the experiments. We use a batch-size of 128 in all the experiments. ADAM optimizer was used in training the joint embedding networks \cite{kingma2014adam}. The model was evaluated on the validation set on the video-text retrieval task after every epoch. To deal with the over-fitting issue, we choose the best model based on the highest sum of recalls.

\subsection{Quantitative Results} \label{sec:quantitative}
We report the experimental results on Charades-STA dataset \cite{gao2017tall} in Table~\ref{tab:charades} and DiDeMo dataset \cite{hendricks2017localizing} in Table~\ref{tab:didemo}.

\vspace{-0.8mm}
\subsubsection{\textbf{Charades-STA Dataset}} 
\vspace{-0.5mm}

The quantitative results on Charades-STA dataset \cite{gao2017tall} are reported in Table~\ref{tab:charades}. The evaluation setup in Charades-STA dataset~\cite{gao2017tall} considers a set of IoU (Intersection over Union) thresholds. We report for IoU $ 0.3, 0.5$ and $0.7$ in Table~\ref{tab:charades}. For these IoU thresholds, we report the recalls - R@1, R@5, and R@10 in Table~\ref{tab:charades}. Following~\cite{gao2017tall}, we use sliding windows of 128 and 256 to obtain the possible temporal segments. The segments are ranked based on the corresponding Text-Guided Attention score.

\vspace{0.05cm}
\textbf{Compared Methods.} We compare our approach with state-of-the-art text to video moment retrieval approaches, CTRL\cite{gao2017tall}, EFRC\cite{xu2018text}, and baseline approaches, VSA-RNN\cite{karpathy2015deep} and VSA-STV\cite{kiros2015skip}. For these methods, we directly cite performances from respective papers when available \cite{gao2017tall, xu2018text}. We report score for VSA-RNN and VSA-STV from \cite{gao2017tall}. If the score for multiple models is reported, we select the score of the best performing method in R@1. Here, VSA-RNN (Visual-Semantic Embedding with LSTM) and VSA-STV (Visual-Semantic Embedding with Skip-thought vector) are text-based image/video retrieval baselines. We also report results for ``Random" which selects a candidate moment randomly. Similar to these approaches, we also utilize the C3D model for obtaining feature representation of videos for fair comparison. We follow the evaluation criteria utilized in \cite{gao2017tall,xu2018text}. 

\vspace{0.03cm}
\textbf{Analysis of Results.} We observe that the proposed approach consistently perform comparably to several fully-supervised approaches in all evaluation metrics. Our weakly-supervised TGA based approach performs significantly better than supervised visual-semantic embedding based approaches VSA-RNN and VSA-STV. We observe that the proposed method achieves a minimum absolute improvement of $6.6\%$ in R@5 and $2.6\%$ in R@1 from VSA-RNN. The maximum relative performance improvement over VSA-STV is $19.3\%$ in R@1 and $13.7\%$ in R@5. We also observe that the proposed approach achieves comparable performance to state-of-the-art method CTRL \cite{gao2017tall} on R@5 evaluation metrics. The proposed approach also shows reasonable performance compared to the EFRC approach \cite{xu2018text}.

\vspace{-0.1cm}
\subsubsection{\textbf{DiDeMo Datset}}
\vspace{-0.05cm}

Table~\ref{tab:didemo} summarizes the results on the DiDeMo dataset \cite{hendricks2017localizing}. DiDeMo only has a coarse annotation of moments. As the videos are trimmed at 30 seconds and the videos are divided into 5-second segments, each video has 21 possible moments. We follow the evaluation setup in \cite{hendricks2017localizing}, which is designed for evaluating 21 possible moments from sentence descriptions. Average of Text-Guided Attention scores of corresponding segments is used as the confidence score for the moments and used for ranking. Following previous works \cite{hendricks2017localizing, xu2018text}, the performance in the dataset is evaluated based on R@1, R@5, and mean intersection over union (mIoU) criteria.

\vspace{0.04cm}
\textbf{Compared Methods.} In Table~\ref{tab:didemo}, we report results for several baselines to analyze the performance of our proposed approach. We divide the table into 3 rows (2.1-2.3). In row-2.1, we report the results of trivial baselines (i.e., Random and Upper-Bound) following evaluation protocol reported in \cite{hendricks2017localizing}. In row-2.2, we group the results of LSTM-RGB-Local \cite{hendricks2017localizing}, EFRC \cite{xu2018text}, and our proposed approach for a fair comparison, as these methods are trained with only the VGG-16 RGB feature. We report the performance of the proposed approach in both validation and test set as LSTM-RGB-local model has been evaluated on validation set \cite{hendricks2017localizing}.
In row-2.3, we report results for state-of-the-art approaches MCN \cite{hendricks2017localizing} and TGN \cite{chen2018temporally}. We also report results of CCA \cite{klein2015associating} and natural language object retrieval based baseline Txt-Obj-Retrieval \cite{hu2016natural} in row-2.3. These methods additionally use optical flow feature along with VGG16 RGB feature. We report the performance of MCN~\cite{hendricks2017localizing}, TGN~\cite{chen2018temporally} and EFRC~\cite{xu2018text} from the respective papers. The results of LSTM-RGB-Local, Txt-Obj-Retrieval, Random, and Upper-Bound are reported from \cite{hendricks2017localizing}. 

\begin{table}[t]
\vspace{0.1cm}
\small
	\caption{This table reports results on the DiDeMo dataset, following the evaluation protocol in \cite{hendricks2017localizing}. Our \textbf{proposed approach} performs on par with several competitive  fully-supervised approaches}
	\vspace{0.1cm}
	\centering
	\begin{tabular}{p{0.4cm}| l|p{0.99cm} p{0.9cm} p{0.9cm}}
		\hline
	\#	& Method        & R@1  & R@5  & mIoU \\
		\hhline{=|=|===}
	\multirow{2}{*}{\ref{tab:didemo}.1}	 & Upper Bound        & 74.75 & 100 & 96.05 \\
		
		 & Random        & 3.75 & 22.5 & 22.64 \\
		\hline 
		
		 \multirow{4}{*}{\ref{tab:didemo}.2} & LSTM-RGB-Local \cite{hendricks2017localizing}        & 13.10 & 44.82 & 25.13  \\

		 & EFRC \cite{xu2018text}  & 13.23 & 46.98 & 27.57  \\
		
		 & \textbf{Proposed (Val. Set)}     & \textbf{11.18} & \textbf{35.62} & \textbf{24.47}  \\
		 
		  & \textbf{Proposed (Test Set)}     & \textbf{12.19} & \textbf{39.74} & \textbf{24.92}  \\
		\hline	
		\multirow{4}{*}{\ref{tab:didemo}.3} & CCA        & 18.11 & 52.11 & 37.82 \\
		 & Txt-Obj-Retrieval \cite{hu2016natural}        & 16.20 & 43.94 & 27.18  \\
		 
		 & MCN \cite{hendricks2017localizing}   & 27.57 & 79.69 & 41.70  \\
		
		& TGN \cite{chen2018temporally}   & 28.23 & 79.26 & 42.97  \\
		\hline

	\end{tabular}
\label{tab:didemo}
\vspace{-0.3cm}
\end{table}

\vspace{0.03cm}
\textbf{Analysis of Results.} Similar to the results on Charades-STA, it is evident from Table~\ref{tab:didemo} that our proposed weakly supervised approach consistently shows comparable performance to several fully-supervised approaches. From row-2.2, we observe that our proposed approach achieves similar performance as LSTM-RGB-Local \cite{hendricks2017localizing} and EFRC \cite{xu2018text}. We observe that R@5 accuracy is slightly lower for our approach compared to supervised approaches. However, R@1 accuracy and mIoU is almost similar. Comparing row-2.3, we observe that the performance is comparable to CCA and Txt-Obj-Retrieval baselines. The performance is low compared to MCN  \cite{hendricks2017localizing} and TGN~\cite{chen2018temporally}. Both of the approaches use additional optical flow features in their framework. MCN additionally use a moment-context feature. Hence, a performance drop is not unexpected. However, we have already observed from the row-2.2 that the performance of our weakly supervised approach is comparable to the MCN baseline model of LSTM-RGB-Local which uses the same RGB feature in training as our method.

\begin{figure*}[t]
\centering
	\includegraphics[width=0.95\textwidth]{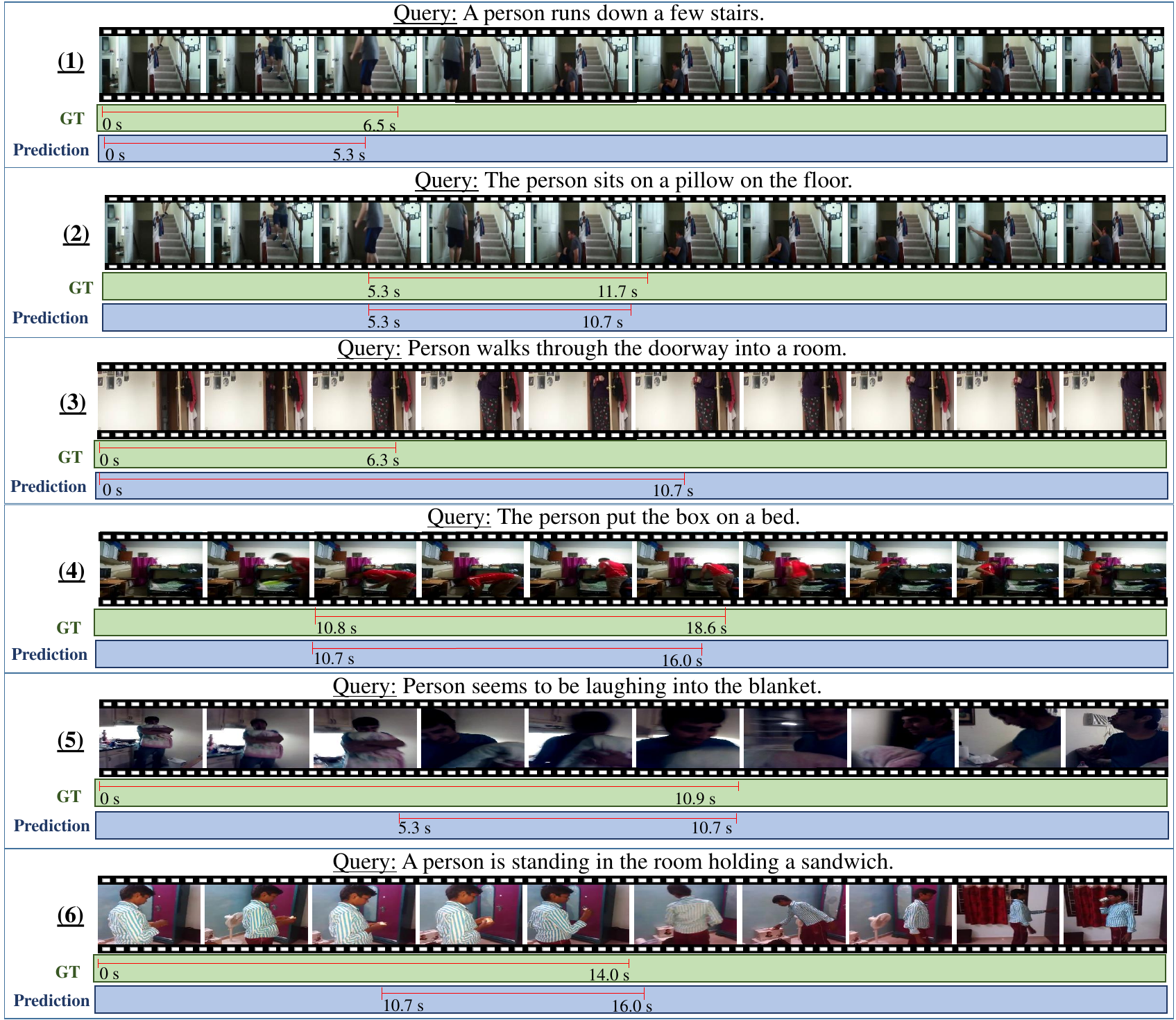}
	\caption{A snapshot of six queries and test videos from Charades-STA dataset with success and failure cases. GT is a ground-truth moment annotation and Prediction is the moment predicted by the proposed weakly-supervised approach. Queries 1, 2, and 4 show success cases where our approach was successful in retrieving the ground truth moment with very high temporal intersection over union (IoU). However, queries 3, 5, and 6 show three cases where our approach was not successful in retrieving the ground truth moment with high IoU. }
	\label{fig:qualit}
	\vspace{-0.3cm}
\end{figure*}

\subsection{Qualitative Results} We provide six qualitative examples of moments predicted by the proposed approach from Charades-STA dataset \cite{gao2017tall} in Fig.~\ref{fig:qualit}. In Fig.~\ref{fig:qualit}, case 1, 2, and 4 show some examples where our approach was successful in retrieving the ground truth moment with high IoU. Cases 1 and 2 are examples where the same video has been used to retrieve different moments based on two different text descriptions. We see our text-aware attention module was successful in finding the correct segment of the video in both the cases. 

While our method retrieves the correct moment from sentence description many cases, it fails to retrieve the correct moment in some cases (e.g., case 3, 5, and 6). Among these three cases, case 3 presents an ambiguous query where the person stands on the doorway but does not enter into the room. The GT moment covers a smaller segment, while our system predicts a longer one. We observe the performance of our system suffers when important visual contents occupy only small portions in frames, e.g., case 5 and 6. In case 6, a sandwich is mentioned in the query which occupies a small portion of frames initially and our framework shifted the start time of the moment to a much later time instant than in the ground truth. Similarly, in case 5, our system was only successful in identifying the person laughing into a blanket after the scene is zoomed in. We believe these are difficult to capture without additional spatial attention modeling or generating region proposals. Moreover, utilizing more cues from videos (e.g., audio, and context) may be helpful in reducing ambiguity in these cases. We leave these as future work.

\section{Conclusions} \label{conclusion}

In this paper, we introduce the novel problem of weakly supervised text to video moment retrieval. In the weakly supervised paradigm, as we do not have access to the temporal boundaries associated with a sentence description, we utilize an attention mechanism to learn the same using only video-level sentences.  Our formulation of the task makes it more realistic compared to existing methods in the literature which require supervision as temporal boundaries or temporal ordering of the sentences. Moreover, the weak nature of the task allows it to learn from easily available web data, which requires minimal effort to acquire compared to manual annotations. Experiments demonstrate that our method in spite of being weakly supervised performs comparably to several fully supervised methods in the literature.


\vspace{0.2cm}
\noindent \textbf{Acknowledgements. } This work was partially supported by NSF grant 1544969 and ONR contract N00014-15-C5113 through a sub-contract from Mayachitra Inc.


{\small
\bibliographystyle{ieee}
\bibliography{egbib}
}

\end{document}